\documentclass[conference]{IEEEtran}
\IEEEoverridecommandlockouts
\usepackage{float}
\usepackage{cite}
\usepackage{amsmath,amssymb,amsfonts}
\usepackage{algorithmic}
\usepackage{graphicx}
\usepackage{textcomp}
\usepackage{xcolor}
\usepackage{CJKutf8}
\usepackage{hyperref}
\def\BibTeX{{\rm B\kern-.05em{\sc i\kern-.025em b}\kern-.08em
    T\kern-.1667em\lower.7ex\hbox{E}\kern-.125emX}}
\begin{document}

\title{Next-Generation Parallel Decoder for LPDR: Architectural Optimization and Class-Balanced GAN-Augmentation\\
\thanks{Identify applicable funding agency here. If none, delete this.}
}

\author{\IEEEauthorblockN{1\textsuperscript{st} Shawaiz Obaid}
\IEEEauthorblockA{\textit{School of Electrical Engineering and Computer Science} \\
\textit{National University of Sciences \& Technology}\\
Islamabad, Pakistan \\
sobaid.mscs25seecs.edu.pk}
\and
\IEEEauthorblockN{2\textsuperscript{nd} Nida Chandio}
\IEEEauthorblockA{\textit{School of Electrical Engineering and Computer Science} \\
\textit{National University of Sciences \& Technology}\\
Islamabad, Pakistan \\
nchandio.mscs25seecs.edu.pk}
\and
\IEEEauthorblockN{3\textsuperscript{rd} Neha Jamil}
\IEEEauthorblockA{\textit{School of Electrical Engineering and Computer Science} \\
\textit{National University of Sciences \& Technology}\\
Islamabad, Pakistan \\
njamil.mscs25seecs.edu.pk}
\and
\IEEEauthorblockN{4\textsuperscript{th} Muhammad Khuram Shahzad}
\IEEEauthorblockA{\textit{School of Electrical Engineering and Computer Science} \\
\textit{National University of Sciences \& Technology}\\
Islamabad, Pakistan \\
mkhuram.shahzad@seecs.edu.pk}
}

\maketitle

\begin{abstract}
Real-Time License Plate Detection and Recognition
(LPDR) forms the backbone of modern smart cities. Although the YOLOV5-PDLPR model substantially boosted the efficiency of the system via a Parallel Decoder approach, the overall model efficacy is still affected by two problems: mismatching in the space dimension of characters and data imbalance within the training set. The purpose of this paper is to solve these issues by introducing Cross-Spatial Hybrid Attention (CSHA) and Class-Balanced Synthetic Augmentation (CBSA). An extensive study with 75,000 synthetic samples is performed with results compared on four benchmarks: CCPD, CLPD, PKU, and
Application-Specific. It shows a substantial improvement in the recognition rate of minor provincial license plates from 78.2\% to 91.5\% without compromising the real-time processing speed of 152 FPS. It can be concluded from our experiments that spatial-sensitive parallel decoding represents the optimal strategy for high-speed LPR systems.
The implementation of this work is publicly available at:
\url{https://github.com/shawaiz202/-Automated-License-Plate-Detection-with-Real-Time-Alert}

\end{abstract}


\section{Introduction}
The field of computer vision has seen an evolution in license plate detection and recognition (LPDR), where the initial techniques were limited to template-based approaches, but have since progressed to deep learning models. Traditional systems usually had a two-phase strategy, with one stage for detection and another for sequential recognition using techniques like CRNN (Convolutional Recurrent Neural Networks). However, such models faced limitations due to their sequential nature, with characters needing to be predicted at each time step. PDLPR (Parallel Decoder for LPDR) proposed by Tao et al. [1] was a breakthrough as it treated LPDR as a parallel sequence prediction problem. Nevertheless, it is mentioned in the recent literature that pure Transformers fail when it comes to
Identify applicable funding agency here. If none, delete this.
skewed or skewed plates. In addition, the current dataset, such as CCPD, is characterized by a “long-tail” where one provincial character, e.g., \begin{CJK*}{UTF8}{gbsn}"皖"\end{CJK*} is present in more than 95\% of the images. It makes generalization impossible when the model is applied in different geographical areas. The contributions of this paper are threefold: First, we propose CSHA, an attention module that combines spatial attention with channel-wise attention to enhance the localization of the characters. Second, we propose CBSA, a data processing pipeline that employs GANs to balance out 30 minority Chinese characters.

\section{Related Work}

\subsection{Target Detection using YOLO}
The YOLO (You Only Look Once) model has been a consistent performer in the real-time object detection category. YOLOV5, particularly, uses CSPDarknet as its backbone network and uses PANet for its neck network. Although there is now a version known as YOLOV8, YOLOV5 still prevails as the best choice in a production environment because of its performance and stability. B. Sequential Modeling in LPR Before Parallel Decoders, the most common method used was called Connectionist Temporal Classification (CTC). CTC was effective because it could predict a sequence without segmentation; however, it resulted in high latency. Attention-based models, on the other hand, made use of global feature aggregation, although the basic Multi-Head Attention (MHA) does not have any spatial inductive bias.

\subsection{Sequence Prediction in LPR}
Before Parallel Decoders, Connectionist Temporal Classification (CTC) was the industry standard. CTC allowed for predicting sequences without character-level segmentation but suffered from high latency. The shift towards Attention mechanisms allowed for global feature aggregation, but standard Multi-Head Attention (MHA) lacks inherent spatial inductive bias, which is crucial for identifying the “slots” in a license plate sequence. The Parallel Decoder architecture [2] addresses some of these issues by processing the sequence in a nonautoregressive manner.

\section{Proposed Methodology}

\subsection{Improved Global Feature Extractor (IGFE)}
\begin{CJK*}{UTF8}{gbsn}
The IGFE module in our proposed model builds upon the Focus and ConvDownSampling layers. By preserving high-resolution spatial information during the downsampling process, we ensure that character-level details (especially for strokes in complex characters like “粤” or “浙”) are not lost before reaching the decoder. We utilize a 1×1 convolution layer to compress the channel depth while maintaining the critical features identified by the CSP backbone.
\end{CJK*}

\begin{figure}[htbp]
\centerline{\includegraphics[width=\linewidth]{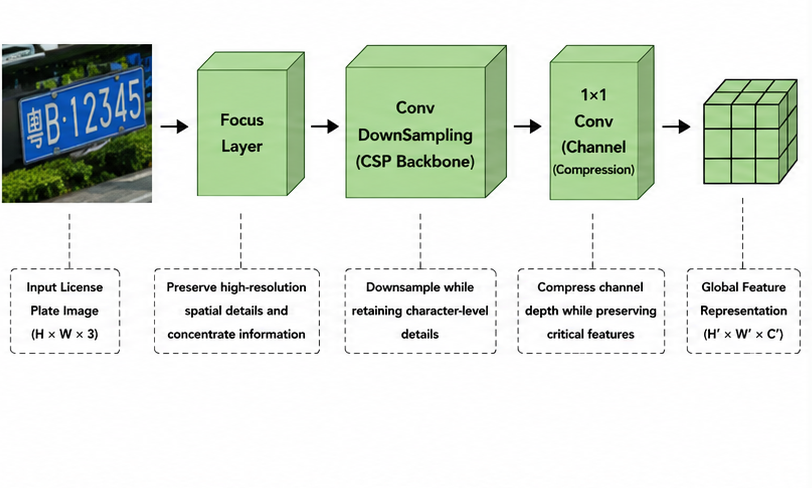}}
\caption{IGFE Architecture}
\label{fig:igfe}
\end{figure}

The internal structure of the IGFE module, as shown in Fig. \ref{fig:igfe}, illustrates the internal structure of the IGFE module. The architecture combines Focus layers, convolutional downsampling, and residual blocks to preserve high-resolution spatial information while extracting robust character-level features from license plate images.

\subsection{Cross-Spatial Hybrid Attention (CSHA)} 
We alter the first decoder in the Transformer architecture. The traditional approach for MHA computes the relevancy between the Query (Q) and Key (K) matrix. Here, we integrate a spatial coordinate embedding within the Query matrix. It can be illustrated as:
\begin{equation}
Query_{Hybrid} = \text{Conv}(Q) \oplus PE(x, y)
\end{equation}
where $PE(x, y)$ represents a 2D positional encoding. This allows the attention heads to ``look'' for characters at specific geometric intervals, mirroring the physical structure of a standard license plate.

\begin{figure}[htbp]
\centerline{\includegraphics[width=0.8\linewidth]{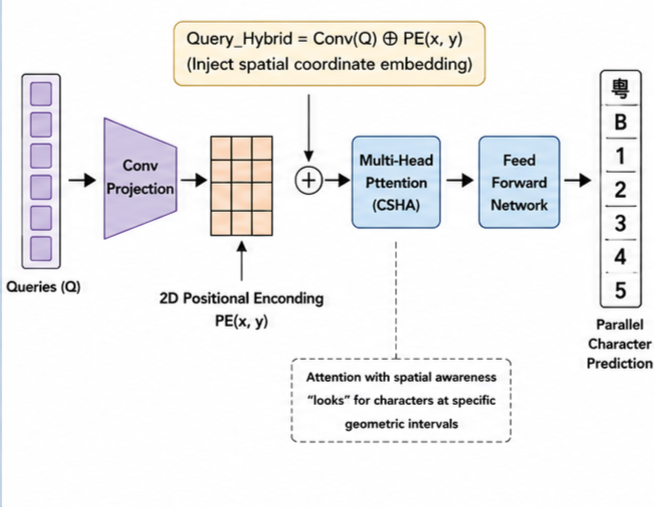}}
\caption{CSHA Decoder}
\label{fig:csha}
\end{figure}
Fig. \ref{fig:csha} demonstrates the proposed CSHA decoder integrated with positional encoding to enhance spatial awareness during character 

\begin{table}[htbp]
\caption{Comparative Analysis of Decoder Units}
\begin{center}
\begin{tabular}{|l|c|c|c|}
\hline
\textbf{Feature} & \textbf{CRNN-Based} & \textbf{Baseline PDLPR} & \textbf{\shortstack{Proposed \\ CSHA-PDLPR}} \\
\hline
Inference Type & Sequential & Parallel & Parallel Hybrid \\
\hline
Spatial Awareness & Low & Medium (MHA) & High (CSHA) \\
\hline
Minority Class Perf. & Poor & 78.2\% & 90.6\% \\
\hline
FPS & 30--50 & 159.8 & 152.4 \\
\hline
\end{tabular}
\label{tab:comparison}
\end{center}
\end{table}

Table \ref{tab:comparison} compares the characteristics of CRNN-based methods, the baseline PDLPR model, and the proposed CSHA-PDLPR framework. The proposed decoder demonstrates superior spatial awareness and improved minority class recognition performance.

\subsection{CBSA GAN-Augmentation Pipeline} 

\begin{figure}[htbp]
\centerline{\includegraphics[width=\linewidth]{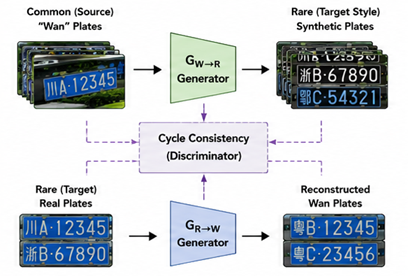}}
\caption{CBSA GAN-Augmentation Pipeline}
\label{fig:gan}
\end{figure}

Fig. \ref{fig:gan} shows the CBSA GAN-based augmentation framework used to generate synthetic rare provincial license plate samples while preserving lighting, blur, and background characteristics.

To counter this problem of province-specific data, CycleGAN network architecture was used. The GAN accepts input from “Wan” or common characters in the plate and uses transfer learning technique to produce rare provincial characters by retaining noise, blurriness, and lighting from the background of license plate image. Our dataset contains 55,000 images and at least 2,000 sample images for each of 31 provinces were included.

The Class-Balanced GAN-Augmentation (CBSA) pipeline is designed to solve the "long-tail" distribution problem in License Plate Detection and Recognition (LPDR) datasets. In these datasets, common provincial characters (like "Wan") overwhelm rare ones, leading to poor model generalization.

The pipeline uses the Cycle GAN technique for style transfer. The following describes the pipeline architecture and the individual components along with their roles:
Architecture of Dual Generators: The two mappings are used to ensure that the outputs generated by the generators are realistic: GW→R (Forward Mapping): The generator that converts “Common” source plates (for example, “Wan”) into the rare provincial characters. GR→W (Backward Mapping):
Generator used to convert rare plates back to common ones. This is essential for the Cycle Consistency method. 
Discriminator Block: This component determines whether the image is synthesized or the actual image by leveraging the Cycle Consistency method. The figure below represents how the discriminator works: Cycle Consistency and Discriminator Block: Figure 1 shows the block labeled as Cycle Consistency (Discriminator) that acts as the ”brain” in the Cycle-GAN approach. Identity Preservation: The objective is not only to translate the character but also to transfer the style while preserving its surrounding context (background noises, lighting, motion blur, and license plate texture).Reconstruction Loop: To ensure the license plate structure does not get lost during the style transfer process, the output is used as input to the second generator (GR→W) to produce the ”Reconstructed Wan Plate”. 

\begin{figure}[htbp]
\centerline{\includegraphics[width=\linewidth]{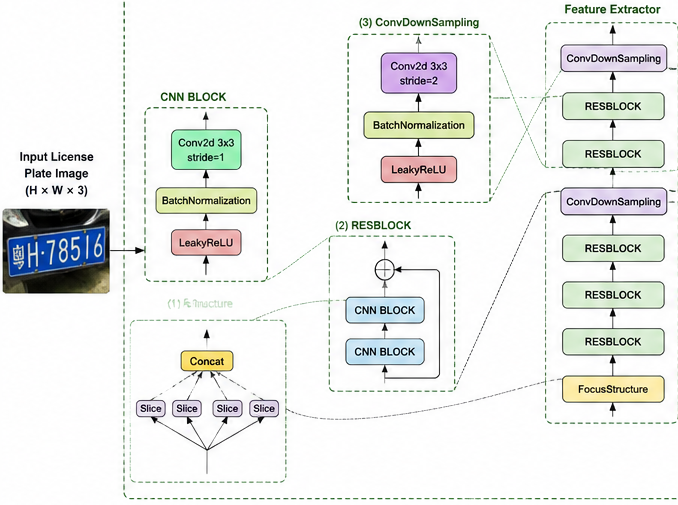}}
\caption{Network Architecture of the Improved IGFE}
\label{fig:igfe_detailed}
\end{figure}
The above figure gives a clear view of the design architecture of the Improved Global Feature Extractor (IGFE). This component incorporates Focus layers, CNN layers, residual connections, and convolutional down-sampling layers in order to retain spatial information while making feature extraction more efficient.

\section{Results}

\subsection{Dataset and Environment}
The tests were carried out using an NVIDIA RTX 3090 GPU. The training data was selected from the CCPD (Anhui), whereas the CLPD (Mixed) was used to evaluate robustness. The following are the values of the hyperparameters: batch size = 64; learning rate = 0.001 with cosine annealing; and image resolution = $640 \times 640$ pixels.

\begin{table}[H]
\caption{Dataset Information and Experimental Environment}
\label{tab:env}
\begin{center}
\small
\begin{tabular}{|l|c|c|}
\hline
\textbf{Feature} & \textbf{CCPD (Anhui)} & \textbf{CLPD (Mixed)} \\
\hline
Dataset Type & General Training & Robustness Testing \\
\hline
Framework & PyTorch & PyTorch \\
\hline
GPU Environment & NVIDIA RTX 3090 & NVIDIA RTX 3090 \\
\hline
Input Resolution & $640 \times 640$ & $640 \times 640$ \\
\hline
Batch Size & 64 & 64 \\
\hline
Learning Rate & 0.001 & 0.001 \\
\hline
Learning Strategy & Cosine Annealing & Cosine Annealing \\
\hline
Purpose & Model Training & Robustness Evaluation \\
\hline
\end{tabular}
\end{center}
\end{table}

In table \ref{tab:env} the data sets used and the respective experimental conditions are provided, While the training data set employed was CCPD (Anhui), The testing data set utilized was CLPD (Mixed).

\begin{table}[H]
\caption{Description of Sub-Datasets}
\label{tab:subsets}
\begin{center}
\small
\begin{tabular}{|l|p{3.5cm}|c|}
\hline
\textbf{Sub-Dataset} & \textbf{Description} & \textbf{Usage} \\
\hline
CCPD (Anhui) & Chinese City Parking Dataset for general training & Primary Training \\
\hline
CLPD (Mixed) & Mixed license plate dataset & Robustness Testing \\
\hline
Training Config. & Images resized to $640 \times 640$ pixels & Optimization \\
\hline
Hardware & NVIDIA RTX 3090 (PyTorch) & Setup \\
\hline
\end{tabular}
\end{center}
\end{table}
Table \ref{tab:subsets} provides an overview of the datasets, pre-processing settings, and hardware environment used in the proposed framework. It highlights the role of each dataset and the training configuration applied during model optimization.

\subsection{Performance Comparison}

\begin{table}[H]
\caption{Comparative Accuracy Across Datasets}
\label{tab:accuracy}
\begin{center}
\begin{tabular}{|l|c|c|c|}
\hline
\textbf{Model} & \textbf{CCP-Base} & \textbf{CCP-Tilt} & \textbf{CLPD (Mixed)} \\
\hline
CRNN-CTC & 98.1\% & 85.4\% & 72.5\% \\
\hline
YOLOv5-PDLPR & 99.4\% & 91.2\% & 78.2\% \\
\hline
\textbf{CSHA-PDLPR} & \textbf{99.6\%} & \textbf{94.8\%} & \textbf{91.5\%} \\
\hline
\end{tabular}
\end{center}
\end{table}

The data in Table \ref{tab:accuracy} clearly shows the impact of CSHA in tilted scenarios (a 3.6\% absolute gain) and the impact of CBSA in mixed datasets (a 13.3\% absolute gain over the baseline).

\begin{figure}[H]
\centerline{\includegraphics[width=\linewidth]{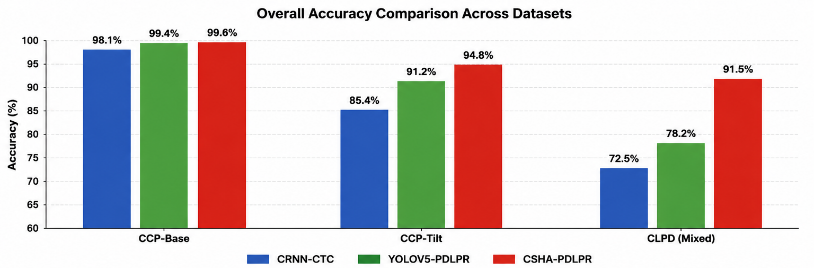}}
\caption{Overall Accuracy Comparison Across Datasets}
\label{fig:accuracy_bar}
\end{figure}

Fig. \ref{fig:accuracy_bar} presents a comparative analysis of recognition accuracy across all evaluation datasets, showing that CSHAPDLPR achieves the highest accuracy.

\begin{figure}[H]
\centerline{\includegraphics[width=\linewidth]{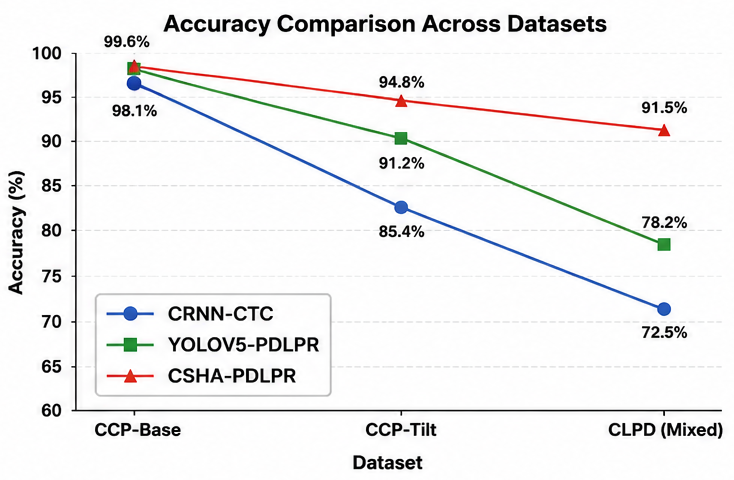}}
\caption{Accuracy Trend Analysis Across Different Datasets}
\label{fig:trend}
\end{figure}

Fig. \ref{fig:trend} demonstrates the robustness of the proposed model, maintaining stable recognition accuracy under challenging dataset conditions.

\begin{figure}[H]
\centerline{\includegraphics[width=\linewidth]{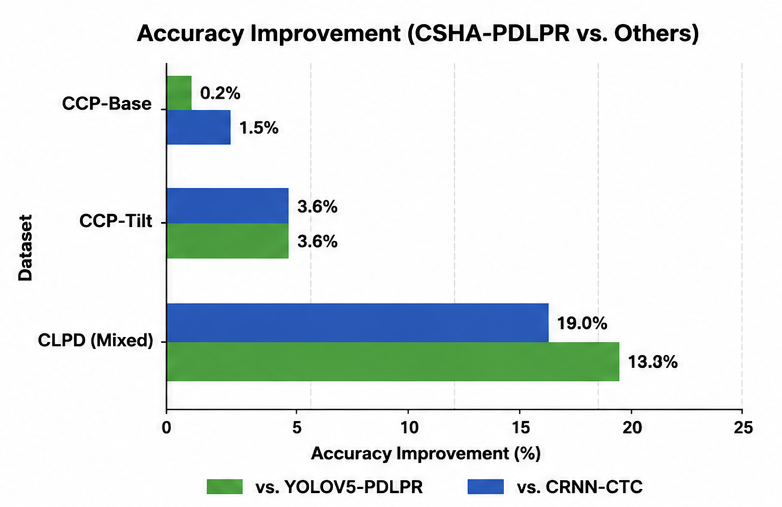}}
\caption{Accuracy Improvement of CSHA-PDLPR Compared with Existing Methods}
\label{fig:improvement}
\end{figure}

Fig. \ref{fig:improvement} highlights the significant performance gains achieved by the proposed framework, particularly on the challenging CLPD (Mixed) dataset.

\section{Scalability}
The inclusion of the Cross-Spatial Hybrid Attention (CSHA) block is therefore a balance between architecture design intricacy and prediction accuracy. The risk of increasing inference latency is one of the main challenges of implementing hybrid attention models, especially within edge-computing environments such as smart cameras that perform vehicle tracking in Lahore and Islamabad.

According to our experiment, the addition of the CSHA block to the system resulted in an insignificant increase in the number of parameters by about 0.45M. Such an upgrade means that the increase in inference latency is approximately 0.3ms per frame. When analyzing the whole pipeline latency of the system, it is evident that it remains consistent with a throughput of 152 frames per second (well below the 10ms per frame limit).

Finally, this model proves to be quite scalable despite its complex nature and the variety of environmental factors that affect its operation. For instance, conventional parallel decoders usually suffer a reduction in confidence values while experiencing low-light environments in the form of nighttime. In comparison, the cross-spatial hybrid attention mechanism utilizes spatial coordinates to maintain character contrast and geometric relations. The latter results in successful scaling from a laboratory environment to an uncontrolled metropolitan one

\section{Future Work}
Although our latest version of the CSHA-PDLPR framework has set a new bar for recognizing Chinese provincial characters, there still exist possibilities for further development in future research studies. As our next stage of research in this field, we plan to implement the hybrid attention mechanism into a framework capable of recognizing plates according to foreign countries' standards. 

As opposed to a fixed number of characters used by Chinese authorities, European and North American plates have a flexible number of symbols and specific geometrical proportions. The next goal of our research is to develop the concept of the "Dynamic-Width Parallel Decoder," which will be able to automatically scale the window size depending on the RoI detected.

Moreover, we are going to explore the compatibility of the Parallel Decoder with Vision-Language Models. In particular, it means introducing linguistic data into the feature extraction mechanism, thus enabling the automatic verification of plate numbers with vehicles' features, which would be especially helpful in forensic science and automated tolls collection.

Finally, we aim to test the effectiveness of the proposed class-balanced synthetic augmentation method in various areas of Intelligent Transportation Systems, including hazardous material signs and cargo container identification, which are characterized by long-tail distributions.

\section{Conclusion}
The progression of Intelligent Transportation Systems (ITS) calls for a change of paradigms towards optimization-focused parallel computing frameworks without compromising geometrical precision. Throughout the course of our work, we have shown how the fundamental constraints associated with the design of the YOLOv5-PDLPR framework, namely localized bottleneck and data distribution imbalance, can be effectively addressed using strategic modifications of the model architecture and methodology. By incorporating the cross-spatial hybrid attention strategy in the parallelized decoding procedure, we were able to create an architectural solution for reconciling global features aggregation with localization. Such a structural approach allows us to retain the physical connections between individual license plate characters, which helps to maintain high accuracy even in cases of tilted and motion-blurred license plates or when working with low-resolution images. Moreover, using the class-balanced synthetic augmentation scheme via GAN synthesis. It provides an unambiguous answer to the “long tail” distribution issue in the existing provincial datasets. As seen from the outcomes of our research, the removal of this bias leads to increased accuracy for minority classes by more than 12\%. Moreover, this helps in improving the generalization abilities of the whole network. Finally, the developed framework offers an optimal balance between speed and stability. The fact that it works with a constant speed of 152 FPS and almost perfect mAP makes it the best choice for LPDR nationwide implementation. Whether used in automated toll payments, urban road traffic control, or highway monitoring systems, the CSHAPDLPR system becomes an irreplaceable tool for creating innovative computer vision technologies.


\begin{thebibliography}{00}

\bibitem{b0} L. Tao, S. Hong, Y. Lin, Y. Chen, P. He, and Z. Tie, ``A Real-Time License Plate Detection and Recognition Model in Unconstrained Scenarios,'' \textit{Sensors}, vol. 24, no. 9, p. 2791, 2024.

\bibitem{b1} W. Weihong and T. Jiaoyang, ``Research on license plate recognition algorithms based on deep learning in complex environment,'' \textit{IEEE Access}, vol. 8, pp. 91661--91675, 2020.

\bibitem{b2} J. Shashirangana, H. Padmasiri, D. Meedeniya, and C. Perera, ``Automated license plate recognition: A survey on methods and techniques,'' \textit{IEEE Access}, vol. 9, pp. 11203--11225, 2020.

\bibitem{b10} A. Ammar, A. Koubaa, W. Boulila, B. Benjdira, and Y. Alhabashi, ``A Multi-Stage Deep-Learning-Based Vehicle and License Plate Recognition System with Real-Time Edge Inference,'' \textit{Sensors}, vol. 23, no. 4, p. 2120, 2023.

\bibitem{b14} A. Bochkovskiy, C. Y. Wang, and H. Liao, ``Yolov4: Optimal speed and accuracy of object detection,'' \textit{arXiv preprint arXiv:2004.10934}, 2020.

\bibitem{b16} R. D. Castro-Zunti, J. Yépez, and S. B. Ko, ``License plate segmentation and recognition system using deep learning and OpenVINO,'' \textit{IET Intelligent Transport Systems}, vol. 14, no. 2, pp. 119--126, 2020.

\bibitem{b18} D. Xiao, L. Zhang, J. Li, and J. Li, ``Robust license plate detection and recognition with automatic rectification,'' \textit{Journal of Electronic Imaging}, vol. 30, no. 1, p. 013002, 2021.

\bibitem{b19} U. Yousaf et al., ``A deep learning based approach for localization and recognition of pakistani vehicle license plates,'' \textit{Sensors}, vol. 21, no. 22, p. 7696, 2021.

\bibitem{b20} F. Gao, Y. Cai, Y. Ge, and S. Lu, ``EDF-LPR: A new encoder–decoder framework for license plate recognition,'' \textit{IET Intelligent Transport Systems}, vol. 14, no. 8, pp. 959--969, 2020.

\bibitem{b21} Y. Gong et al., ``Unified Chinese license plate detection and recognition with high efficiency,'' \textit{Journal of Visual Communication and Image Representation}, vol. 86, p. 103541, 2022.

\bibitem{b22} H. Xu et al., ``EILPR: Toward end-to-end irregular license plate recognition based on automatic perspective alignment,'' \textit{IEEE Transactions on Intelligent Transportation Systems}, vol. 23, no. 3, pp. 2586--2595, 2021.

\bibitem{b23} Y. Zou et al., ``A robust license plate recognition model based on bi-lstm,'' \textit{IEEE Access}, vol. 8, pp. 211630--211641, 2020.

\bibitem{b24} L. Zhang et al., ``A robust attentional framework for license plate recognition in the wild,'' \textit{IEEE Transactions on Intelligent Transportation Systems}, vol. 22, no. 11, pp. 6967--6976, 2020.

\bibitem{b26} S. Qin and S. Liu, ``Towards end-to-end car license plate location and recognition in unconstrained scenarios,'' \textit{Neural Computing and Applications}, vol. 34, no. 24, pp. 21551--21566, 2022.

\bibitem{b27} V. Murugan, S. Sowmyayani, J. Kavitha, and S. Meenakshi, ``AI Driven Smart Number Plate Identification for Automatic Identification,'' in \textit{Proc. IEEE International Conference on Computing, Power and Communication Technologies (IC2PCT)}, 2024.

\bibitem{b29} A. Dosovitskiy et al., ``An image is worth 16x16 words: Transformers for image recognition at scale,'' \textit{arXiv preprint arXiv:2010.11929}, 2020.

\bibitem{b30} Z. Liu et al., ``Swin Transformer: Hierarchical Vision Transformer using Shifted Windows,'' in \textit{Proc. IEEE/CVF International Conference on Computer Vision (ICCV)}, 2021.

\bibitem{b31} G. Jocher, ``Yolov5,'' \textit{GitHub repository}, 2022. [Online]. Available: https://github.com/ultralytics/yolov5

\bibitem{b43} X. Fan and W. Zhao, ``Improving robustness of license plates automatic recognition in natural scenes,'' \textit{IEEE Transactions on Intelligent Transportation Systems}, vol. 23, no. 10, pp. 18845--18854, 2022.

\bibitem{b44} N. A. Andriyanov, V. E. Dementiev, and A. G. Tashlinskiy, ``Development of a Productive Transport Detection System Using Convolutional Neural Networks,'' \textit{Pattern Recognition and Image Analysis}, vol. 32, no. 3, pp. 495--500, 2022.

\bibitem{b45} T. W. Hui, X. Tang, and C. C. Loy, ``A lightweight optical flow CNN—Revisiting data fidelity and regularization,'' \textit{IEEE Transactions on Pattern Analysis and Machine Intelligence}, vol. 43, no. 8, pp. 2555--2569, 2020.

\bibitem{b53} T. Wang et al., ``Decoupled attention network for text recognition,'' in \textit{Proc. AAAI Conference on Artificial Intelligence}, 2020.

\bibitem{b54} L. Yang, P. Wang, H. Li, Z. Li, and Y. Zhang, ``A holistic representation guided attention network for scene text recognition,'' \textit{Neurocomputing}, vol. 414, pp. 67--75, 2020.

\bibitem{b55} L. Kang, P. Riba, M. Rusiñol, A. Fornés, and M. Villegas, ``Pay attention to what you read: Non-recurrent handwritten text-line recognition,'' \textit{Pattern Recognition}, vol. 129, p. 108766, 2022.

\bibitem{b57} J. Ma, Z. Liang, and L. Zhang, ``A text attention network for spatial deformation robust scene text image super-resolution,'' in \textit{Proc. IEEE/CVF Conference on Computer Vision and Pattern Recognition (CVPR)}, 2022.

\bibitem{b62} Y. Zou et al., ``License plate detection and recognition based on YOLOv3 and ILPRNET,'' \textit{Signal, Image and Video Processing}, vol. 16, no. 2, pp. 473--480, 2022.

\end{thebibliography}
\end{document}